\documentclass[10pt,twocolumn,letterpaper]{article}

\usepackage[pagenumbers]{cvpr} 

\definecolor{cvprblue}{rgb}{0.21,0.49,0.74}
\usepackage[breaklinks,colorlinks,allcolors=cvprblue]{hyperref}
\usepackage{graphicx}
\usepackage[export]{adjustbox}
\usepackage{algorithm}
\usepackage{algpseudocode}
\usepackage{makecell}
\usepackage{xcolor}
\usepackage{array}
\usepackage[strings]{underscore}
\usepackage[symbol]{footmisc}
\usepackage[utf8]{inputenc}
\usepackage{tikz}
\usetikzlibrary{positioning}
\usepackage{booktabs}
\usepackage[accsupp]{axessibility}
\usepackage{multirow}
\usepackage{tabularx}

\newcommand{\setimgdir}[1]{\graphicspath{{#1/}}}

\newcommand{\simpleimg}[1]{%
    \adjustbox{frame}{%
        \includegraphics[width=\linewidth]{#1}%
    }%
}

\newcommand{\overlaysize}{0.25\linewidth}

\newcommand{\overlayimg}[2]{%
  \makebox[\linewidth][l]{%
    \begin{tikzpicture}
      \node[inner sep=0pt] (main) {%
        \adjustbox{frame}{%
          \includegraphics[width=0.90\linewidth]{#1}%
        }%
      };

      \node[
        anchor=south east,
        inner sep=0pt,
        xshift=0.10\linewidth,
        yshift=-0.21\linewidth
      ] at (main.north east) {%
        \adjustbox{frame}{%
          \includegraphics[
            width=\overlaysize,
            height=\overlaysize
          ]{#2}%
        }%
      };
    \end{tikzpicture}%
  }%
}

\title{Hist2Style: Histogram-Guided Stylization with Bilateral Grids}

\stepcounter{footnote}
\author{
Dekel Galor$^{1,2,}\thanks{Work partially done during an internship at Adobe.}$ \quad 
Adam Pikielny$^{1}$ \quad 
Zhoutong Zhang$^{1}$ \quad 
Ke Wang$^{1}$ \\ 
Laura Waller$^{2}$ \quad 
Jiawen Chen$^{1}$ \quad 
Ilya Chugunov$^{1}$ \\\\
$^{1}$Adobe Nextcam \quad $^{2}$University of California, Berkeley \\
}
\begin{document}
\maketitle

\begin{abstract}

Photorealistic style transfer aims to match the color and tone of an input image to that of a style target while preserving the content and details of the original scene. Although existing large image models can facilitate these kinds of appearance edits, their high computational demands, potential for hallucinations, and limited user control make them unsuitable for high-resolution, real-time workflows. We introduce Hist2Style, a bilateral-grid formulation for fast, edge-aware stylization that preserves visual fidelity by constraining operations to locally affine transforms in bilateral space. Our model distills a large image editing model into a lightweight network by training on a large supervised corpus generated with language and vision-language models, targeting spatially varying color edits. The network conditions on a histogram-based embedding of the style target to provide an interpretable interface for adjusting the output style by modifying the target color distribution. Overall, Hist2Style maintains content structure by construction, avoids hallucinations, and supports real-time, high-resolution photorealistic stylization with interactive user-controllable color and tone adjustments. Our project page is available at \href{https://dgalor.github.io/hist2style/}{dgalor.github.io/hist2style/}.

\end{abstract}

\section{Introduction}
\label{sec:intro}

Color and tone define much of an image’s mood, coherence, and narrative~\cite{colorPsych}, and photographers and filmmakers have long relied on color grading to evoke emotion and maintain visual consistency across scenes~\cite{colorPsych, movieGrading, haine2019color}. Yet this process remains painstakingly manual, requiring individual adjustment of lighting, contrast, and palette across images and video clips~\cite{haine2019color}. Automating such color grading while preserving realism motivates the study of style transfer: the ability to borrow the visual \textit{look} of one image and apply it to another without altering its content~\cite{nstAreview, luan2017deep, Gatys15}.

While the goal of style transfer is conceptually simple, what constitutes \textit{style} and \textit{content} varies considerably across methods. In this work, we adopt the photorealistic convention: style is defined by color and tone, content by structure and edges~\cite{luan2017deep, ReinhardTransfer2001}. Many existing approaches instead rely on deep neural features that, while expressive, are difficult to interpret or control~\cite{nstAreview}, leaving users with little flexibility to achieve their intended look~\cite{luan2017deep}.

\begin{figure}[t]
    \centering
    \includegraphics[width=\linewidth]{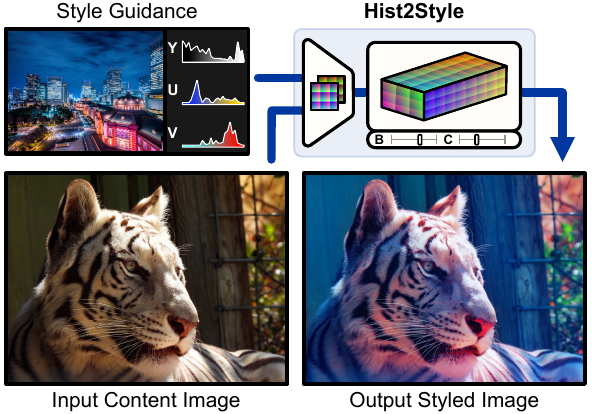}
    \caption{Our method preserves content and detail by mapping nearby pixels of similar color to similar outputs via a lightweight network conditioned on a histogram-based style embedding. This enables fast, high-resolution stylization, and the histogram itself is directly editable, giving users intuitive control over color and tone.}
    \label{fig:teaser}
    \vspace{-1.5em}
\end{figure}

\begin{figure*}[t]
    \centering
    \includegraphics[width=0.85\textwidth]{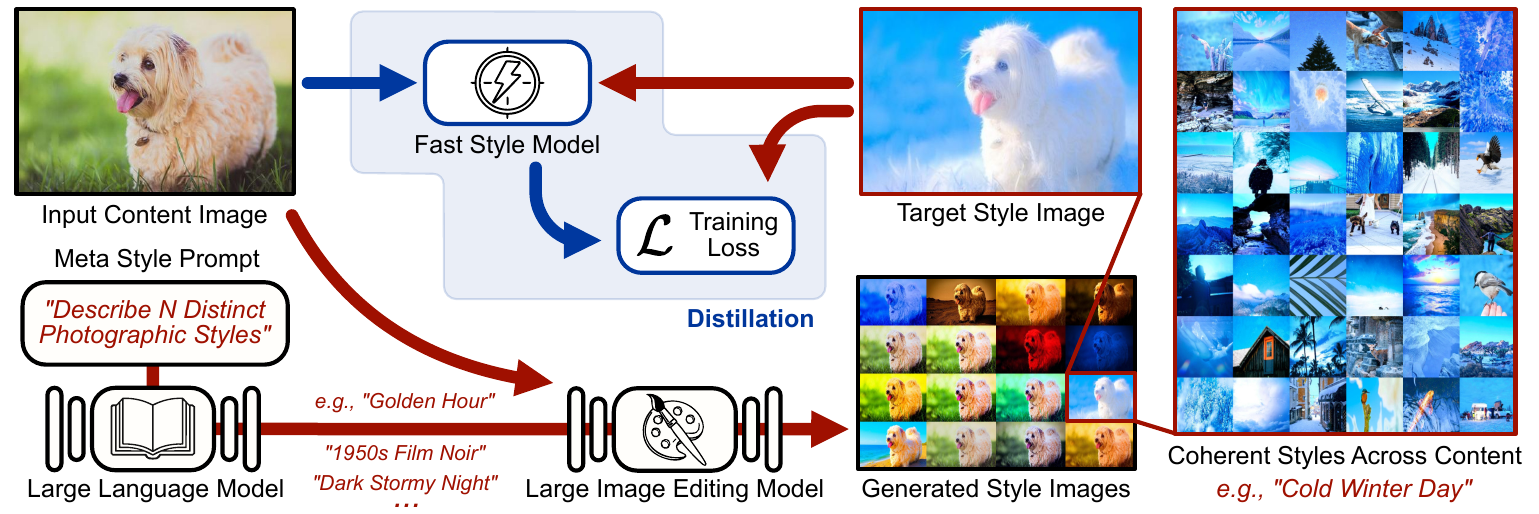}
    \caption{
        \textbf{Selective distillation} to condense the capabilities of a large image editing model into a lightweight, specialized network for photorealistic stylization. We procedurally generate style editing prompts with a large language model. The prompts are used to instruct an editing model to edit standard images into different style variations, which are coherent across images. Our model is then trained to mimic the stylization of the large editing model with a regression loss (\textit{red:} precomputed, \textit{blue:} live training).
    }
    \label{fig:dataset}
\end{figure*}

Recently, large image editing models have transformed how users approach stylization~\cite{imageEditingSurvey, relightingImageEditingSurvey, hertz2022prompt}. These models let users specify visual intentions through image or text prompts, promising far greater flexibility than traditional style transfer methods~\cite{hertz2022prompt}. In principle, their expressive power could render specialized photorealistic stylization algorithms obsolete, but in practice this versatility introduces key limitations.
\textit{Performance:} To support diverse editing tasks, these models sacrifice efficiency for generality, demanding high computation, large memory, and long inference times~\cite{zhang2025styletransferdecadesurvey}. \textit{Hallucination:} Edits can introduce identity drift, structural distortions, and other artifacts that break photorealism~\cite{mokady2022nulltextinversioneditingreal}. \textit{Controllability:} Expressing subtle color and tone through prompts introduces ambiguities that hinder stylistic precision~\cite{Brooks2023}.

We address these limitations by selectively distilling a large image editing model into an efficient sub-model specialized for photorealistic stylization. To improve \textit{performance}, the distillation compresses a foundation model into a lightweight task-specific network. To suppress \textit{hallucinations}, we constrain edits to locally affine transformations in bilateral space. To enhance \textit{controllability}, we introduce a histogram-based style embedding that enables intuitive, interactive manipulation of color and tone. Named Hist2Style, our model foregoes content generation to focus strictly on photorealism: precisely adjusting color and tone without distorting the image structure.

Our key contributions are:
\begin{enumerate}
    \item A photorealistic, histogram-guided stylization network robust to hallucinations and scalable to high resolutions.
    \item An interactive system for real-time image editing with artistically expressive, histogram-based control.
    \item A stylization quality assessment metric for automated and reproducible evaluation of stylization models that correlates highly with human preference.
\end{enumerate}

\section{Related Works}
\begin{figure*}[t]
    \centering
    \includegraphics[width=0.95\textwidth]{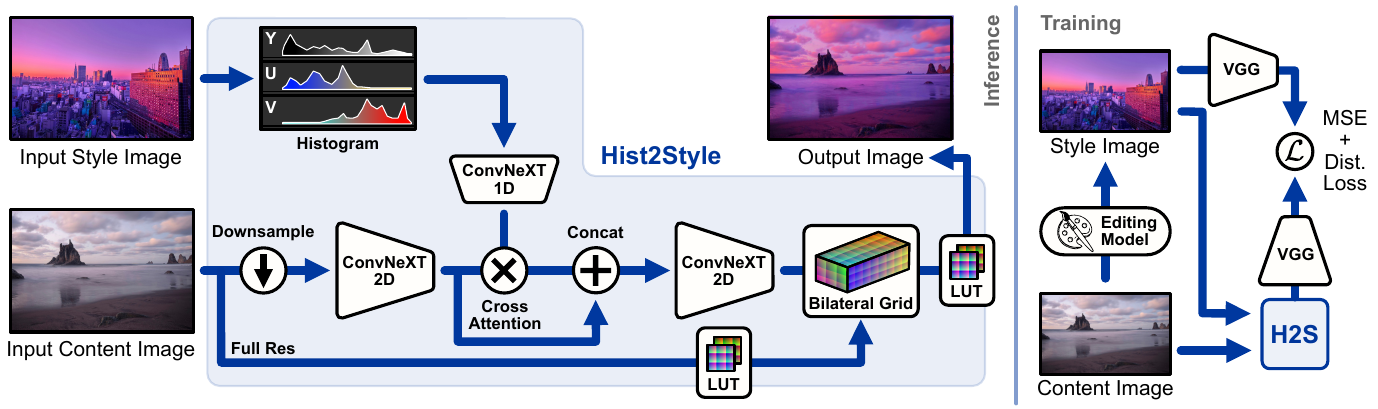}
    \caption{
        \textbf{Model architecture (left)}. Our model separately embeds a downsampled content image and style color histogram via ConvNeXt blocks. The two are then fused with cross attention, and fed through the output head to produce a spatially adaptive color transform known as the bilateral grid. The bilateral grid is applied to the content image and fed through a learned per-channel nonlinearity (LUT) to produce a stylized image. \textbf{Training (right)} is done by first creating synthetic ground-truth stylization pairs using a large image-editing model. Then the model is fed a content image, the histogram of the ground-truth image, and trained to regress to the ground-truth with a perceptual loss.
    }
    \label{fig:method}
\end{figure*}

The field of image style transfer has evolved through several distinct paradigms, each defining \textit{style} and \textit{content} differently and balancing trade-offs between semantic richness, photorealism, and user control~\cite{nstAreview, zhang2025styletransferdecadesurvey}.

\vspace{2pt}\noindent\textbf{Classical Color Transfer}
Early methods of image stylization focused on transferring the global color distribution, adjusting a photograph’s overall color statistics to match a reference~\cite{ReinhardTransfer2001, Pitie05a, Pitie05b, Pitie07a, Pitie07b, Pitie08}. 
Approaches from the early 2000s such as Reinhard \etal~\cite{ReinhardTransfer2001} matched mean and variance of color channels between images, while Pitié \etal~\cite{Pitie05a, Pitie05b, Pitie07a, Pitie07b, Pitie08} introduced more advanced methods such as \textit{iterative distribution transfer} (IDT)~\cite{Pitie07a} to align full histograms.
These techniques could be categorized as \textit{photorealistic} since they preserve the original spatial structure~\cite{ReinhardTransfer2001, Pitie05a, Pitie05b, Pitie07a, Pitie07b, Pitie08}. However, as these color mappings are calculated from and applied to global color statistics, they lack spatial or semantic awareness~\cite{SALUT}. This property often leads to visible artifacts. For example, \cite{Pitie07b} reports ``graininess'' when source and target images differ greatly and requires postprocessing to suppress. Our work shares the spirit of these classical color transfer methods, but extends them with semantic and spatial awareness.

\vspace{2pt}\noindent\textbf{Neural Style Transfer}
A significant change in the field was marked by the rise of neural networks, which redefined style transfer~\cite{nstAreview}. Gatys \etal introduced the seminal method for artistic style transfer, defining style not as a global color profile but as the second-order statistics (Gram matrices) of deep features from a VGG network~\cite{Gatys15, vgg}. This unlocked unprecedented stylistic richness but was inherently ``artistic'' and non-photorealistic~\cite{Gatys15, luan2017deep}. Subsequent works like Universal Style Transfer (WCT) accelerated this process to interactive speeds, but remained focused on artistic application, whereas our objective is photorealistic stylization~\cite{WCT, huang2017adain}.
Achieving photorealistic style transfer became a major focus around 2017~\cite{zhang2025styletransferdecadesurvey, luan2017deep}. Luan \etal introduced Deep Photo Style Transfer, which added a photorealism regularizer and used semantic segmentation to constrain Gatys’ style loss~\cite{luan2017deep, Gatys15}. This avoided the “painting” distortions and kept stylized results plausible, but the method required slow per-image optimization~\cite{luan2017deep}. Li \etal proposed PhotoWCT, which applied a similar constraint but as a postprocessing step for WCT~\cite{PhotoWCT, WCT}. Yoo \etal alleviated the need for postprocessing with WCT\textsuperscript{2}, a wavelet-based stylization network that preserves image structures~\cite{WCT2}. Afifi \etal applied histogram style conditioning and guided upsampling to reduce artifacts with ReHistoGAN \cite{histGAN}. Another significant breakthrough came from Xia \etal~\cite{Xia2020}, which reframed the soft photorealism regularization as learning to predict the parameters of local affine color transforms known as bilateral grids~\cite{Xia2020, Gharbi_2017}. Subsequent works continued to improve efficiency and fidelity. PhotoWCT\textsuperscript{2} achieved more efficient inference while maintaining the quality of PhotoWCT~\cite{PhotoWCT2}. D-LUT employed a diffusion process for super-efficient color distribution transfer, but constrained to global adjustments~\cite{DLUT}. SA-LUT then introduced spatially varying LUTs via quadrilinear interpolation~\cite{SALUT}.  These methods all share a strategy of constraining the transform space to prevent the free-form distortions of artistic style transfer, thereby producing outputs that could pass as real photographs~\cite{zhang2025styletransferdecadesurvey}. Such constraints appear in various forms, including wavelet filters~\cite{WCT2, PhotoWCT2}, local color maps~\cite{luan2017deep, Xia2020, SALUT}, and global color maps~\cite{DLUT, nlut, NeuralPreset}. We opt for using the bilateral grid, a local color map used in~\cite{Xia2020}, since it is efficient, scalable, and provides a balance between expressivity and photorealistic constraints~\cite{Xia2020, Gharbi_2017, Adams2010, Barron2015A, BarronPoole2016, wang2024bilateralguidedradiancefield}.

\vspace{2pt}\noindent\textbf{Foundation Models for Image Editing}
A new paradigm has since emerged with the rise of foundational image editing models~\cite{imageEditingSurvey, relightingImageEditingSurvey, Brooks2023, mokady2022nulltextinversioneditingreal, FluxKontext, wu2025qwenimagetechnicalreport}. Scaling trends in diffusion and transformer architectures have enabled general-purpose systems capable of performing a wide range of edits~\cite{Brooks2023, mokady2022nulltextinversioneditingreal}, from recoloring~\cite{relightingImageEditingSurvey} and relighting~\cite{relightingImageEditingSurvey} to object replacement~\cite{relightingImageEditingSurvey}, all within a single unified model~\cite{relightingImageEditingSurvey}. Open-source models such as Flux Kontext~\cite{FluxKontext} and Qwen Image Edit~\cite{wu2025qwenimagetechnicalreport} exemplify this shift toward universal editing~\cite{relightingImageEditingSurvey}. These systems achieve remarkable flexibility, reframing visual manipulation as a form of conditional generation rather than task-specific transformation~\cite{imageEditingSurvey}. Yet this generality introduces new challenges: high computational demands~\cite{zhang2025styletransferdecadesurvey}, unpredictable hallucinations~\cite{mokady2022nulltextinversioneditingreal}, and limited user control over precise outcomes~\cite{Brooks2023}.

In this work, we leverage the semantic priors and expressivity of image editing models while avoiding the hallucination~\cite{mokady2022nulltextinversioneditingreal} and performance~\cite{zhang2025styletransferdecadesurvey} issues by sacrificing generality (adding photorealistic constraints) and distilling into a lightweight model. Additionally, we tackle the controllability~\cite{Brooks2023} issue by building an interactive mechanism for matching results to user intent.

\section{Hist2Style}

We train a lightweight neural network that maps a content image and a style embedding to a stylized image as illustrated in \cref{fig:method}. We define the style embedding as the marginal (per channel) histogram of the style image, providing interpretability and controllability (discussed further in \cref{sec:methods:histogram} and \cref{sec:methods:interactive}).

\subsection{Selective Distillation of Image Editing Models}
\label{sec:methods:distillation}

Our network is trained to distill a subset of the capabilities of a large image editing model~\cite{FluxKontext}. To this end, we first prompt a large language model~\cite{gemini} to generate a diverse set of names and descriptions of photorealistic styles such as ``Golden Hour'', ``1950s Film Noir'', etc. Then we take a standard photography dataset~\cite{UnsplashLiteDataset} and prompt the image editing model~\cite{FluxKontext} with the LLM-generated descriptions to create many stylized versions of each image in the dataset.
As discussed in \cref{sec:intro}, large image editing models struggle with hallucinations~\cite{mokady2022nulltextinversioneditingreal} and controllability issues~\cite{Brooks2023}, which affect some images generated with the teacher model. We therefore filter teacher outputs by VGG19 feature cosine similarity, retaining pairs above 0.5 (1.1M of 1.7M generated edits), and propose the use of bilateral grids to bake photorealistic constraints into the model itself~\cite{Gharbi_2017, Xia2020}.

\subsection{Bilateral Grids for Photorealistic Stylization}
\label{sec:methods:bilateral}

The bilateral grid is a parameterization of image-to-image functions that explicitly encodes the edges of 2D images via a guidance dimension. By storing affine transformations at each grid cell, the bilateral grid can compactly represent locally affine edge-preserving functions, a desirable property for photorealistic stylization~\cite{luan2017deep,Xia2020}. The bilateral grid decouples the resolution of the transform from that of the image: computing the transform is done by \emph{slicing} into the grid using multilinear interpolation followed by a per-pixel operation. This lets our method scale to arbitrarily high resolutions.
In this work, we adopt the strategy from prior methods and define the guidance dimension of the bilateral grid as a learned function of RGB $g: \mathbb{R}^3 \rightarrow \mathbb{R}$~\cite{Gharbi_2017,Xia2020}. This effectively acts as a luminance dimension that prevents smoothing across edges.
For content image $I_c$ with shape $(H, W, 3)$, the model predicts a bilateral grid of size $(G_g, G_h, G_w)$, corresponding to the grid resolution in the guidance dimension, height, and width, respectively. We use a grid size of $(8, 16, 16)$, which proved effective in~\cite{Xia2020}. For each grid entry, the model predicts an affine transform, which corresponds to $3 \times 4 = 12$ scalar values. Additionally, the model predicts a premultiplied $\alpha$ value per grid entry as a measure of model uncertainty. The result of slicing the bilateral grid via trilinear interpolation is a per-pixel affine transform, which is applied to the content image to produce the stylized output.

\subsection{Histogram-Guided Style Transfer}
\label{sec:methods:histogram}

Stylization is often encoded via features of pretrained image models~\cite{luan2017deep, huang2017adain, Xia2020} or language prompt embeddings~\cite{relightingImageEditingSurvey}. However, these representations lack two critical properties: \textit{interpretability}, as disentangling style from content within a neural network is non-trivial~\cite{porquet2025copyingstyleextractingvalue}; and \textit{controllability}, since matching network embeddings to a user's precise intent remains challenging~\cite{Brooks2023, porquet2025copyingstyleextractingvalue}.

\noindent To address the limitations in interpretability and control, we represent style using the marginal color histogram of the style image. This choice is motivated by the histogram's foundational role in traditional photo editing~\cite{AdobeLightroom2007}. Unlike classic methods, however, our goal is not pure distribution transfer, which can introduce artifacts by disregarding spatial structure~\cite{Pitie07a}. Instead, we leverage synthetic training data, and jointly optimize spatial and distribution objectives so that the model learns both. 

We use MSE as a spatial loss between the output and ground truth images. For our distribution loss, we estimate the squared one-dimensional Wasserstein-2 metric by sorting pixels across space, and computing the MSE~\cite{peyré2020computationaloptimaltransport} (see \cref{alg:w2_distance}). We find that applying these losses in perceptual space (VGG~\cite{vgg}) yields better generalization than applying them in image color space, an observation we attribute to imperfections in the synthetic dataset.

\begin{algorithm}[t]
\caption{Marginal distribution matching loss.}
\label{alg:w2_distance}
\begin{algorithmic}[1]
\State \textbf{Input:} flattened image features $X, Y \in \mathbb{R}^{HW \times C}$
\State \textbf{Output:} 1D Wasserstein loss $L$
\For{each channel $c \in \{1,...,C\}$}
    \State $X_c = X[:, c]$, $Y_c = Y[:, c] \in \mathbb{R}^{HW}$
    \State Sort: $x_c \gets \mathrm{sort}(X_c)$, \quad $y_c \gets \mathrm{sort}(Y_c)$
    \State Compute squared distance: $d_c = \|x_c - y_c\|^2$
\EndFor
\State \textbf{Return:} mean loss $L = \frac{1}{C} \sum_{c=1}^{C} d_c$
\end{algorithmic}
\end{algorithm}

\subsection{Model Architecture}
As illustrated in \cref{fig:method}, our model follows a dual-branch design that processes the content image and the style embedding separately, then fuses them to predict a spatially adaptive color transform  represented as a bilateral grid~\cite{Xia2020, Gharbi_2017}.

The content branch is a convolutional encoder with ConvNeXt-style blocks introduced by Liu \etal~\cite{convNext}. Each block includes large-kernel depthwise convolutions $(7\times7)$ and inverted bottleneck layers with GELU activations and LayerNorm. The content encoder downsamples the image spatially to match the intended bilateral grid size, $16\times16$ in our experiments. In parallel, the style branch encodes the source and target color histogram. We treat the 1D histograms as a sequence, one per color channel, and apply an analogous ConvNeXt-structured 1D CNN~\cite{convNext}. This yields a global style token representing the style editing operation. 

To inject the global style editing information into the local content features, we employ a cross-attention module~\cite{attention}. We then concatenate the content image features with the results of this module, and feed the combined features into a small convolutional output head that predicts an affine bilateral grid~\cite{Xia2020}.

In addition to the affine coefficients of the grid, the model predicts an uncertainty channel $\alpha$ with a softplus nonlinearity for stability. Next, the affine coefficients get multiplied by $\alpha$ in-place, while the $\alpha$ channel itself remains untouched. The content image is used to \textit{slice} into the grid using trilinear interpolation, resulting in per-pixel affine transform coefficients which are divided by the final \textit{sliced} $\alpha$ values. Before and after applying the affine transforms, we apply an additional per-channel nonlinearity, parameterized as smooth monotonic functions.

{
    \setimgdir{figs/unsplash_imgs2}
    
    \begin{figure*}[!ht]
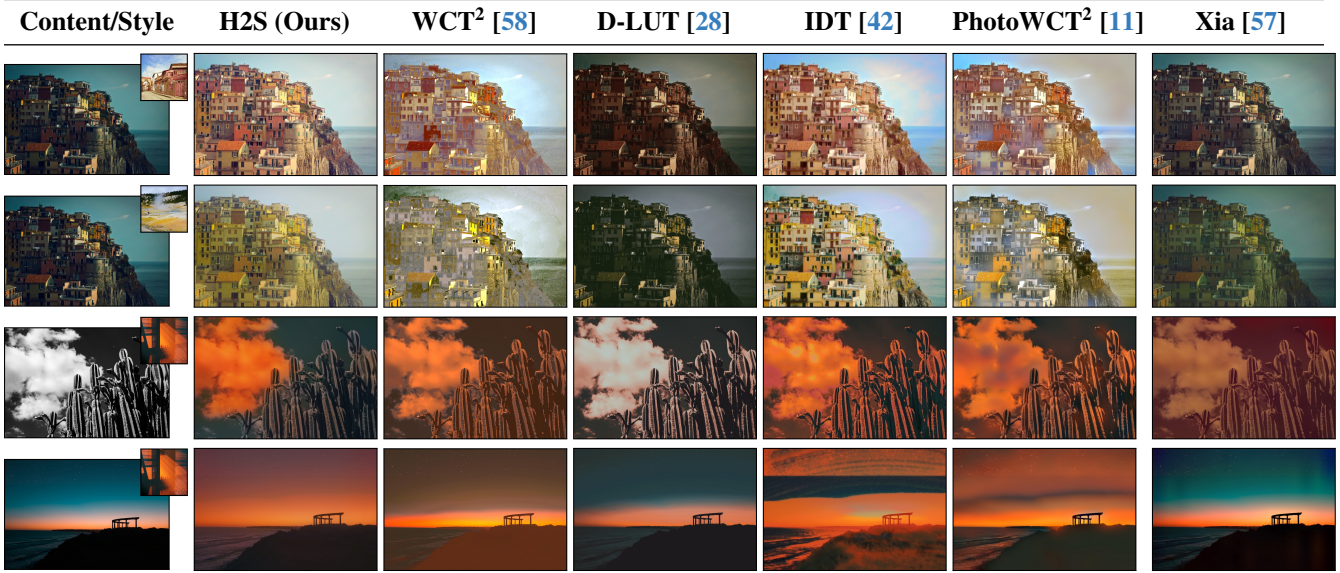

        \centering

        \setlength{\tabcolsep}{1.5pt}
        \begin{tabularx}{\textwidth}{@{} *{7}{X} @{}}
            \toprule
\multicolumn{1}{c}{\textbf{Content/Style}} &
\multicolumn{1}{c}{\textbf{H2S (Ours)}} &
\multicolumn{1}{c}{\textbf{WCT\textsuperscript{2}~\cite{WCT2}}} &
\multicolumn{1}{c}{\textbf{D-LUT~\cite{DLUT}}} &
\multicolumn{1}{c}{\textbf{IDT~\cite{Pitie07a}}} &
\multicolumn{1}{c}{\textbf{PhotoWCT\textsuperscript{2}~\cite{PhotoWCT2}}} &
\multicolumn{1}{c}{\textbf{Xia~\cite{Xia2020}}} \\
            \midrule
            
            \overlayimg{c-c92_s-s13/content.jpg}{c-c92_s-s13/style.jpg}\hspace{0.6em} &
\simpleimg{c-c92_s-s13/H2S.jpg} &
\simpleimg{c-c92_s-s13/WCT2.jpg} &
\simpleimg{c-c92_s-s13/dlut.jpg} &
\simpleimg{c-c92_s-s13/idt.jpg} &
\simpleimg{c-c92_s-s13/photowct2.jpg} &
\simpleimg{c-c92_s-s13/xia.jpg} \\
            
            \overlayimg{c-c92_s-s18/content.jpg}{c-c92_s-s18/style.jpg}\hspace{0.6em} &
\simpleimg{c-c92_s-s18/H2S.jpg} &
\simpleimg{c-c92_s-s18/WCT2.jpg} &
\simpleimg{c-c92_s-s18/dlut.jpg} &
\simpleimg{c-c92_s-s18/idt.jpg} &
\simpleimg{c-c92_s-s18/photowct2.jpg} &
\simpleimg{c-c92_s-s18/xia.jpg} \\
            
            \overlayimg{c-c93_s-s3/content.jpg}{c-c93_s-s3/style.jpg}\hspace{0.6em} &
\simpleimg{c-c93_s-s3/H2S.jpg} &
\simpleimg{c-c93_s-s3/WCT2.jpg} &
\simpleimg{c-c93_s-s3/dlut.jpg} &
\simpleimg{c-c93_s-s3/idt.jpg} &
\simpleimg{c-c93_s-s3/photowct2.jpg} &
\simpleimg{c-c93_s-s3/xia.jpg} \\

            \overlayimg{c-c15_s-s3/content.jpg}{c-c15_s-s3/style.jpg}\hspace{0.6em} &
\simpleimg{c-c15_s-s3/H2S.jpg} &
\simpleimg{c-c15_s-s3/WCT2.jpg} &
\simpleimg{c-c15_s-s3/dlut.jpg} &
\simpleimg{c-c15_s-s3/idt.jpg} &
\simpleimg{c-c15_s-s3/photowct2.jpg} &
\simpleimg{c-c15_s-s3/xia.jpg} \\

        \end{tabularx}
        \vspace{-0.5em}
        \caption{Qualitative comparisons with baseline methods on images from the user-study evaluation set (\cref{tab:study}), illustrating how Hist2Style produces compelling, spatially adaptive stylizations that avoid high-frequency edge and color artifacts (best viewed zoomed in).}
        \label{fig:eval_grid}
    \end{figure*}
}

{
    \setimgdir{figs/ilya_imgs3}
    
    \begin{figure*}[!ht]
        \centering

        \setlength{\tabcolsep}{1.5pt}
        \begin{tabularx}{\textwidth}{@{} *{7}{X} @{}}
            \toprule
\multicolumn{1}{c}{\textbf{Content/Style}} &
\multicolumn{1}{c}{\textbf{H2S (Ours)}} &
\multicolumn{1}{c}{\textbf{WCT\textsuperscript{2}~\cite{WCT2}}} &
\multicolumn{1}{c}{\textbf{D-LUT~\cite{DLUT}}} &
\multicolumn{1}{c}{\textbf{IDT~\cite{Pitie07a}}} &
\multicolumn{1}{c}{\textbf{PhotoWCT\textsuperscript{2}~\cite{PhotoWCT2}}} &
\multicolumn{1}{c}{\textbf{Xia~\cite{Xia2020}}} \\
            \midrule
            
            \overlayimg{c-c13_s-s6/content.jpg}{c-c13_s-s6/style.jpg} &
\simpleimg{c-c13_s-s6/H2S.jpg} &
\simpleimg{c-c13_s-s6/WCT2.jpg} &
\simpleimg{c-c13_s-s6/dlut.jpg} &
\simpleimg{c-c13_s-s6/idt.jpg} &
\simpleimg{c-c13_s-s6/photwct2.jpg} &
\simpleimg{c-c13_s-s6/xia.jpg} \\
            
            \overlayimg{c-c13_s-s8/content.jpg}{c-c13_s-s8/style.jpg} &
\simpleimg{c-c13_s-s8/H2S.jpg} &
\simpleimg{c-c13_s-s8/WCT2.jpg} &
\simpleimg{c-c13_s-s8/dlut.jpg} &
\simpleimg{c-c13_s-s8/idt.jpg} &
\simpleimg{c-c13_s-s8/photwct2.jpg} &
\simpleimg{c-c13_s-s8/xia.jpg} \\
            
            \overlayimg{c-c21_s-s8/content.jpg}{c-c21_s-s8/style.jpg} &
\simpleimg{c-c21_s-s8/H2S.jpg} &
\simpleimg{c-c21_s-s8/WCT2.jpg} &
\simpleimg{c-c21_s-s8/dlut.jpg} &
\simpleimg{c-c21_s-s8/idt.jpg} &
\simpleimg{c-c21_s-s8/photwct2.jpg} &
\simpleimg{c-c21_s-s8/xia.jpg} \\
            
            \overlayimg{c-c23_s-s13/content.jpg}{c-c23_s-s13/style.jpg} &
\simpleimg{c-c23_s-s13/H2S.jpg} &
\simpleimg{c-c23_s-s13/WCT2.jpg} &
\simpleimg{c-c23_s-s13/dlut.jpg} &
\simpleimg{c-c23_s-s13/idt.jpg} &
\simpleimg{c-c23_s-s13/photwct2.jpg} &
\simpleimg{c-c23_s-s13/xia.jpg} \\
            
            \overlayimg{c-c30_s-s12/content.jpg}{c-c30_s-s12/style.jpg} &
\simpleimg{c-c30_s-s12/H2S.jpg} &
\simpleimg{c-c30_s-s12/WCT2.jpg} &
\simpleimg{c-c30_s-s12/dlut.jpg} &
\simpleimg{c-c30_s-s12/idt.jpg} &
\simpleimg{c-c30_s-s12/photwct2.jpg} &
\simpleimg{c-c30_s-s12/xia.jpg} \\
            
            \overlayimg{c-c3_s-s3/content.jpg}{c-c3_s-s3/style.jpg} &
\simpleimg{c-c3_s-s3/H2S.jpg} &
\simpleimg{c-c3_s-s3/WCT2.jpg} &
\simpleimg{c-c3_s-s3/dlut.jpg} &
\simpleimg{c-c3_s-s3/idt.jpg} &
\simpleimg{c-c3_s-s3/photwct2.jpg} &
\simpleimg{c-c3_s-s3/xia.jpg} \\
            
        \end{tabularx}
        \vspace{-0.5em}
                \caption{Qualitative comparisons with baseline methods on an independently collected, non-public dataset, highlighting Hist2Style’s performance across a wide range of style–content pairs, including daytime, evening, vibrant, and monochrome scenes.}
        \label{fig:ilya_grid}
    \end{figure*}
}
\begin{figure*}[t]
    \centering
    \includegraphics[width=\textwidth]{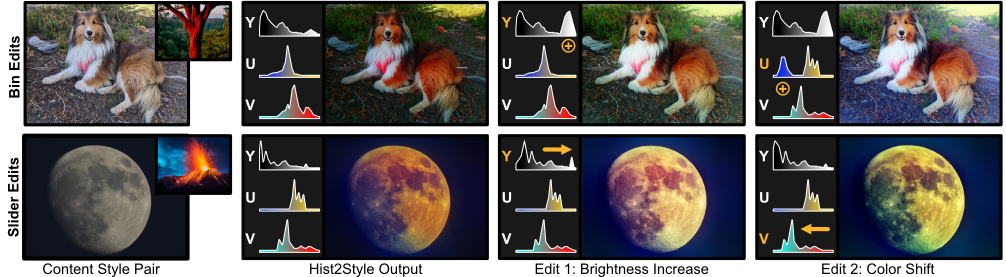}
    \caption{
        \textbf{User control}. We can interactively control the stylization process by directly editing the guiding histogram. This \textit{global} user intent is fed to Hist2Style, which applies \textit{local} changes that are adaptive to the image being edited. Users can edit the histograms directly by dragging the curve or use more familiar operations via custom sliders (see \cref{sec:methods:interactive} for more details).
    }
    \label{fig:editing}
\end{figure*}

\subsection{Implementation}
We implement all models in PyTorch~\cite{pytorch} with PyTorch Lightning~\cite{pytorchLightning, wandb}. Optimization is performed with Adam~\cite{adam} using a learning rate of $3\times10^{-4}$, $(\beta_1,\beta_2)=(0.9,0.99)$, together with a linear warm-up schedule of one epoch. The model has 1.5M parameters, and was trained for 1127 epochs (each of 22.5K images) with batches of 64 images. Experiments are conducted on a single A100 GPU.

During training, for each content image, the dataloader randomly samples two style variants, one used as the input content image, and the other as simultaneously both the style image and ground truth image. Standard data augmentations such as random horizontal flips and resized crops are applied to improve generalization.

\subsection{Interactive Histogram Manipulation}
\label{sec:methods:interactive}

Histogram-conditioned stylization admits a natural interactive interface: rather than adjusting the output directly, the user manipulates the input histogram fed to the network, which translates \textbf{global} intent into \textbf{locally} coherent edits guided by priors learned during training. This mirrors traditional photography workflows such as sliders and tone curves~\cite{AdobeLightroom2007}, while operating entirely in histogram space to give users interpretable, precise control over color and tone. A demonstration video is available on our \href{https://dgalor.github.io/hist2style/}{project page}.

We implement the following sliders for the marginal histogram in Y'CbCr color space:
\begin{itemize}
    \item Exposure slider $E \in [0, 1]$ applies a multiplicative factor on the luminance channel.
    \item Contrast slider $C \in [0, \infty]$ interpolates between a delta function at the peak of the luminance ($C=0$), and the original luminance histogram ($C=1$).
    \item U-shift slider $U \in [-1, 1]$ horizontally shifts the histogram mass of the Cb channel.
    \item V-shift slider $V \in [-1, 1]$ horizontally shifts the histogram mass of the Cr channel.
    \item Smoothing slider $S \in [0, 1]$ applies a smoothing function.
\end{itemize}

Additionally, we include an amount slider  $A \in [-\infty, \infty]$ applied to the output of the model, which interpolates between the identity transformation ($A=0$) and the model's predicted transformation ($A=1$). Apart from slider control, we developed an interface for ``direct histogram manipulation'', allowing users to manually sculpt the histogram by dragging it. This mechanism can selectively add mass to, or subtract it from, a certain luma or chroma bin, while the model ensures the transformation obeys photorealistic image statistics, illustrated in the top row of \cref{fig:editing}.

\section{Results}

The data used for training is synthetically generated from the Unsplash Lite dataset consisting of 25K high-quality images~\cite{UnsplashLiteDataset}. We generated an average of 6\,7 synthetic style variations per image, resized to $256\times256$px for training.

For evaluation, we first randomly selected 200 content photos from the Unsplash dataset~\cite{UnsplashLiteDataset} which we did not use for training, and then manually curated 136 natural images, removing ones with artificial or edited content. We also curated 19 style images from the Unsplash website~\cite{unsplash} that were not in the training set.

We compare our algorithm's automated stylization to the state of the art in photorealistic style transfer via a user study. Then we compare to standard performance metrics such as runtime, memory, cycle consistency, and color matching score. Lastly, we propose an automated VLM-based Stylization Quality Assessment (SQA) metric which we find to be highly aligned with user preference.

Beyond automation, histogram guidance allows for fine user control and creative expression, as shown in \cref{fig:editing}. Further ablations on model components and training are shown in \cref{fig:ablations}. Qualitative comparisons are presented in \cref{fig:eval_grid} and \cref{fig:ilya_grid}.

\begin{table}[t]
\caption{We conducted a \textbf{user study} comparing Hist2Style (H2S) to prior work. The table reports the percentage of trials that H2S won, tied, or lost against each baseline method. Our method outperforms the baselines in this study, winning over 61\% of the trials against any single method. Note that SA-LUT was trained on log-encoded images which may contribute to its lower performance on RGB according to authors.}
\label{tab:study}
\centering
\footnotesize
\begin{tabular}{lccc}
\toprule
Method & H2S Win \% & Tie \% & Lose \% \\
\midrule
SA-LUT & \textbf{82.57\%} & 3.56\% & 13.86\% \\
WCT\textsuperscript{2} & \textbf{72.58\%} & 5.85\% & 21.57\% \\
Xia \etal & \textbf{73.24\%} & 4.10\% & 22.66\% \\
IDT & \textbf{73.75\%} & 3.01\% & 23.25\% \\
D-LUT & \textbf{70.59\%} & 3.04\% & 26.37\% \\
PhotoWCT\textsuperscript{2} & \textbf{61.62\%} & 6.26\% & 32.12\% \\
\bottomrule
\end{tabular}
\end{table}

\begin{table}[t]
\caption{\textbf{Runtime} (s) is reported across image resolutions and models. Hist2Style and Xia \etal are the fastest models when evaluated on novel content-style pairs. D-LUT requires an amortized cost for new style images, but is faster to apply if style is known in advance.}
\label{tab:runtime}
\centering
\footnotesize
\setlength{\tabcolsep}{4pt}
\begin{tabular}{lccccc}
\toprule
Method & $256^2$ & $512^2$ & $1024^2$ & $2048^2$ & $4096^2$ \\
\midrule
\textbf{Hist2Style}      & \textbf{0.001} & \textbf{0.003} & \textit{0.009} & \textit{0.04} & \textit{0.1} \\
\textbf{Xia \etal}       & \textit{0.003} & \textbf{0.003} & \textbf{0.004} & \textbf{0.008} & \textbf{0.03} \\
D-LUT                    & \makecell{100} & \makecell{100} & \makecell{100} & \makecell{100} & \makecell{100} \\
SA-LUT                   & 0.2 & 0.2 & 0.2 & 0.2 & 0.2 \\
PhotoWCT\textsuperscript{2} & 0.3 & 0.3 & 0.3 & 0.4 & 1 \\
IDT                      & 0.1 & 0.2 & 0.3 & 0.4 & 0.9 \\
WCT\textsuperscript{2}   & 0.04 & 0.07 & 0.1 & 0.4 & OOM \\
ReHistoGAN               & 0.01 & 0.08 & 0.47 & 2.22 & 8.83 \\
\bottomrule
\end{tabular}
\end{table}

\subsection*{User Study}

Leveraging the evaluation dataset described above, we designed a two-choice, anonymous user study to assess preference among stylized output. All comparisons used the \textbf{default} output of our model \textit{without} user control, ensuring a fair comparison of automated stylization quality. In each trial, participants were shown a random content-style pair along with two stylized outputs, Hist2Style and a randomly selected baseline presented in random order. See \cref{fig:eval_grid} for example images from the study. Participants were instructed to choose ``the image that best preserves the structure and details of the content image, while matching the color palette, tone, and overall aesthetic of the style image without introducing artifacts''. They could also mark a trial as a tie when no clear preference existed.

We collected 3,000 valid trials from 31 experts in photography. As summarized in \cref{tab:study}, Hist2Style achieves the highest overall preference, winning more than 61\% of all recorded trials. Against PhotoWCT\textsuperscript{2}, the next strongest method, over 6\% of comparisons resulted in ties; after accounting for these, PhotoWCT\textsuperscript{2} attains a win rate below~33\%. In addition to perceptual quality, Hist2Style provides an order-of-magnitude improvement in both runtime and peak memory usage relative to PhotoWCT\textsuperscript{2}, as discussed in \cref{sec:results:runtime}. To facilitate comparison with other metrics, we define the User Score for each baseline method as its win rate against Hist2Style plus half the tie rate $(\text{H2S Lose \%} + \tfrac{1}{2}\text{Tie \%}$) as used in \cref{fig:sqa} and \cref{tab:quantitative_results}.

\newpage
\subsection*{Metrics}
\paragraph{Stylization Quality Assessment}

\begin{figure}[t]
    \centering
    \includegraphics[width=0.95\linewidth]{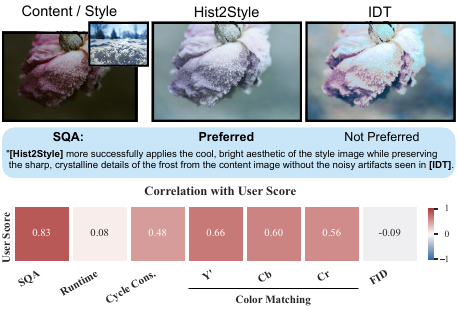}
    \caption{
        \textbf{Stylization Quality Assessment} (SQA) metric. We introduce SQA, a new metric for assessing stylization quality which is aligned to human preference, as shown in the correlation plot. For clarity, metrics were oriented so that higher is better. 
    }
    \label{fig:sqa}
\end{figure}
\begin{table}[t]
\centering
\caption{Quantitative comparison of style transfer methods. Lower is better ($\downarrow$) for cycle consistency, color matching, and FID~\cite{pyiqa,parmar2021cleanfid}, while higher is better ($\uparrow$) for User Score and SQA (N=7000).}
\label{tab:quantitative_results}
\resizebox{\columnwidth}{!}{
\begin{tabular}{@{} l cc c ccc c @{}}
\toprule
\multirow{2}{*}{Method} & \multirow{2}{*}{\begin{tabular}{@{}c@{}}User\\Score ($\uparrow$)\end{tabular}} & \multirow{2}{*}{\begin{tabular}{@{}c@{}}SQA\\($\uparrow$)\end{tabular}} & \multirow{2}{*}{\begin{tabular}{@{}c@{}}Cycle Cons.\\(MSE $\downarrow$)\end{tabular}} & \multicolumn{3}{c}{Color Matching ($W^2_2 \downarrow$)} & \multirow{2}{*}{\begin{tabular}{@{}c@{}}FID\\($\downarrow$)\end{tabular}} \\
\cmidrule(lr){5-7}
& & & & Y' & Cb & Cr & \\
\midrule
Hist2Style & \textbf{50.00} & \textbf{50.00} & \textit{401.92} & 387.30 & 49.85 & 80.53 & 71.44 \\
PhotoWCT\textsuperscript{2} & \textit{35.25} & \textit{36.70} & 1012.90 & \textbf{224.86} & \textbf{23.14} & \textbf{52.41} & 112.07 \\
Xia \etal & 24.71 & 24.12 & 646.40 & 2510.83 & 123.06 & 175.31 & \textbf{50.46} \\
D-LUT & 27.89 & 25.30 & 430.35 & 3850.52 & 201.46 & 307.20 & 87.65 \\
SA-LUT & 15.64 & 28.20 & 1613.97 & 4194.92 & 215.25 & 280.91 & 71.72 \\
IDT & 24.75 & 25.70 & \textbf{215.57} & 2692.25 & 169.78 & 203.85 & 92.32 \\
WCT\textsuperscript{2} & 24.50 & 16.10 & 990.43 & \textit{308.57} & \textit{29.06} & \textit{55.09} & 97.96 \\
ReHistoGAN & - & 33.90 & 583.97 & 3000.75 & 136.84 & 207.18 & \textit{66.84} \\
\bottomrule
\end{tabular}
}
\end{table}

Image stylization is inherently subjective, making quantitative evaluation challenging. While user studies remain the most reliable approach, they are difficult to replicate and their results can be ambiguous to interpret. To address this, we propose a stylization quality assessment (SQA) metric based on a vision-language model (VLM). Following recent work leveraging VLMs for image quality assessment~\cite{machineIQA, afine}, we prompt a large VLM~\cite{gemini} with the same question posed to photographers in our user study. As shown in \cref{fig:sqa}, SQA is highly correlated with the user score. For direct comparison with other metrics, see \cref{tab:quantitative_results}.

\paragraph{Cycle Consistency}

Inspired by~\cite{cycleGan}, we assess methods' \textit{cycle consistency} by stylizing an image from style A to B and back to A, and reporting the MSE relative to the original in \cref{tab:quantitative_results}. IDT attains the lowest cycle-consistency error as it maps colors globally back to their original distribution. Hist2Style follows, reinforcing our claims of spatial consistency, as local affine transformations in one bilateral grid can be effectively reversed by those of a subsequent grid.

\paragraph{Color Matching}
To assess how well each method reproduces the target color statistics, we evaluate the distribution-matching score defined in \cref{alg:w2_distance} on the predicted stylizations. The results are summarized in \cref{tab:quantitative_results}. PhotoWCT\textsuperscript{2} and WCT\textsuperscript{2} achieve the lowest scores, corresponding to a high stylization strength. Hist2Style follows, demonstrating a strong compromise between reliable matching of global color statistics and higher perceptual quality, as supported by the user score and SQA in \cref{tab:quantitative_results}.

\subsubsection*{Runtime, Resolution, and Memory}
\label{sec:results:runtime}

Runtime and memory usage are key considerations for practical deployment of image-editing models, particularly on mobile or edge devices where compute and thermal budgets are limited. We therefore evaluate runtime and memory consumption across a range of input resolutions, with results summarized in \cref{tab:runtime}.

Hist2Style and Xia \etal are the two fastest methods. Hist2Style is faster at lower resolutions, and both methods run in 0.003\,s at $512\times512$. At higher resolutions, Xia \etal scales more favorably, processing 16\,MP images in 0.03\,s compared to 0.1\,s for Hist2Style. All remaining baselines are several times slower. Although D-LUT incurs an amortized initialization cost of 100\,s per style image, its runtime is the lowest (0.001\,s at 16\,MP) once this cost is paid. Hist2Style also remains lightweight in terms of memory usage, requiring only 1\,GB to process 4\,MP images and 5\,GB at 16\,MP.

\begin{figure}[t]
    \centering
    \includegraphics[width=0.95\linewidth]{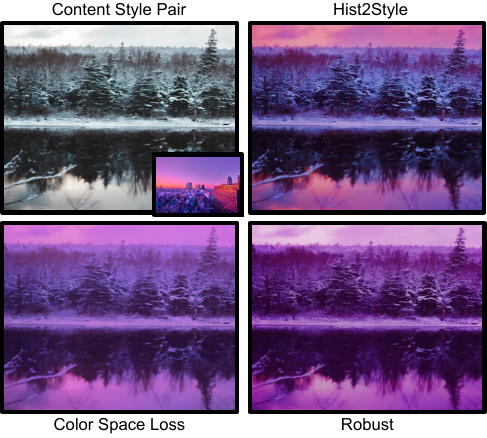}
    \caption{
        \textbf{Ablation studies and extensions.} We explored different architectural and training configurations. "Color Space Loss" refers to the output from a model where the loss was applied to raw color space during training, instead of perceptual space. Compared to this model, Hist2Style expresses a richer color transfer and better maintains image contrast. "Robust" refers to the output from a model trained not with the true histogram of the target image, but with that of another image in the same style. Compared to the robust model, Hist2Style displays similar contrast but improved color richness and accuracy.
    }
    \label{fig:ablations}
\end{figure}

\subsection*{Ablations and Extensions}

\paragraph{Robust Training.} Our training paradigm synthesizes ground truth images that are absent in test time. While our dataset's coherent styles allow us to mimic test-time conditions, the increased ambiguity causes such models to struggle with accurate color matching (see \cref{fig:ablations}).

\paragraph{Color Space Loss.} We find that using a perceptual loss improves fidelity (see \cref{fig:ablations}). We attribute this to imperfect pixel alignment in our artificially generated data.

\section{Discussion and Future Work}
This work presents Hist2Style, a fast photorealistic stylization model that distills spatially varying color edits from a large image editing model into a lightweight network. By constraining the transform to local affine operations in bilateral space and conditioning on a histogram-based style embedding, Hist2Style preserves content and fine details while supporting expressive, spatially adaptive color changes and a simple interactive editing framework. In future work, we hope our proposed model can be further tailored to specific image editing tasks such as:
\label{sec:limitations_futurework}

\paragraph{Multi-Layer Stylization.} Although bilateral grids are spatially varying, a user may want further spatial control, such as assigning different styles to different objects in the scene. This could be explored in future work by co-optimizing a segmentation model to produce masks and interpolating the affine transforms across multiple selected styles and masks.
\paragraph{Nondestructive Content Editing.} Although we focus on stylization without altering content, for some styles it may be necessary to add effects such as film grain, speckle, or bokeh. A potential direction for future work could combine Hist2Style's nondestructive color adjustments with optional effect layers to reproduce these looks while preserving photorealism.
\paragraph{Video and 3D Assets.} While Hist2Style is designed for image stylization, the approach could potentially be extended to other domains such as video~\cite{movieGrading, WCT2, DLUT} and 3D representations~\cite{wang2024bilateralguidedradiancefield}. Future work would include enforcing temporal consistency for long video sequences and integrating the method with radiance-field-based scene representations~\cite{wang2024bilateralguidedradiancefield}.

\section*{Acknowledgments}
We thank Marc Levoy, Florian Kainz, Yifei Fan, Kevin Zhang, Ruiming Cao, Lars Jebe, Ethan Weber, Justin Yu, the Adobe Nextcam team, and WallerLab for valuable discussions and feedback. We thank the creators of software packages~\cite{seabold2010statsmodels, pedregosa2011scikit, reback2020pandas, hunter2007matplotlib, waskom2021seaborn, abid2019gradio}. Dekel Galor is supported by the National Science Foundation Graduate Research Fellowship Program under Grant No.~DGE-1752814, and by the Center for Innovation in Vision and Optics. Laura Waller is a Chan Zuckerberg Biohub SF investigator, and partially funded by U.S. Air Force Office of Scientific Research award no. FA955-22-1-0521.

{
    \small
    \bibliographystyle{ieeenat_fullname}
    \bibliography{main}
}

\end{document}